\documentclass[11pt,a4paper]{article}
\usepackage[table]{xcolor}
\usepackage{emnlp2020_style}
\usepackage{times}
\usepackage{multirow}
\usepackage{latexsym}
\usepackage{booktabs,arydshln}
\usepackage[flushleft]{threeparttable}

\newcommand{\sqlquery}[1]{\texttt{\small #1}}
\usepackage{bigstrut}
\newcommand{\hide}[1]{}

\usepackage{microtype}
\usepackage{natbib}
\usepackage{graphicx}
\usepackage{subfiles}
\usepackage{todonotes}
\usepackage{amsmath}
\usepackage{mathptmx}
\usepackage{xspace}

\aclfinalcopy 

\newcommand{\Ours}{\textsc{ColloQL}\xspace}

\def\|#1|{\mathid{#1}}
\newcommand{\mathid}[1]{\ensuremath{\mathit{#1}}}
\def\<#1>{\codeid{#1}}
\protected\def\codeid#1{\ifmmode{\mbox{\fontsize{9pt}{10.8pt}\selectfont \smaller\ttfamily{#1}}}\else{\fontsize{9pt}{10.8pt}\selectfont \smaller\ttfamily
		#1}\fi}

\title{\Ours: Robust Cross-Domain Text-to-SQL Over Search Queries}

\author{Karthik Radhakrishnan\thanks{\hspace{5pt} This research was conducted during the author's internship at Salesforce.} \\
  Carnegie Mellon University \\
  \texttt{\fontsize{10.3}{13.6}\selectfont kradhak2@cs.cmu.edu} \\\And 
  Arvind Srikantan \\
  Salesforce Inc. \\
  \texttt{\fontsize{10.3}{13.6}\selectfont asrikantan@salesforce.com} \\\And
  Xi Victoria Lin \\
  Salesforce Research \\
  \texttt{\fontsize{10.3}{13.6}\selectfont xilin@salesforce.com}}
\date{}
\makeatletter
\def\thickhline{%
  \noalign{\ifnum0=`}\fi\hrule \@height \thickarrayrulewidth \futurelet
   \reserved@a\@xthickhline}
\def\@xthickhline{\ifx\reserved@a\thickhline
               \vskip\doublerulesep
               \vskip-\thickarrayrulewidth
             \fi
      \ifnum0=`{\fi}}
\makeatother

\newlength{\thickarrayrulewidth}
\begin{document}
\maketitle

\begin{abstract}

Translating natural language utterances to executable queries is a helpful technique in making the vast amount of data stored in relational databases accessible to a wider range of non-tech-savvy end users. Prior work in this area has largely focused on textual input that is linguistically correct and semantically unambiguous. However, real-world user queries are often succinct, colloquial, and noisy, resembling the input of a search engine. In this work, we introduce data augmentation techniques and a sampling-based content-aware BERT model (\Ours) to achieve robust text-to-SQL modeling over natural language search (NLS) questions. Due to the lack of evaluation data, we curate a new dataset of NLS questions and demonstrate the efficacy of our approach. \Ours's superior performance extends to well-formed text, achieving 84.9\% (logical) and 90.7\% (execution) accuracy on the WikiSQL dataset, making it, to the best of our knowledge, the highest performing model that does not use execution guided decoding. 
\end{abstract}
\section{Introduction}
Relational databases store a vast amount of the world's data and are typically accessed via structured query languages like SQL. A natural language interface to these databases (NLIDB) could significantly improve the accessibility of this data by allowing users to retrieve and utilize the information without any programming expertise.
With the release of large-scale datasets \cite{wikisql,DBLP:conf/acl/RadevKZZFRS18,spider}, this task has gained a lot of attention and has been widely studied in recent years.

\begin{figure}[h!]
    \centering
    \includegraphics[scale=0.21]{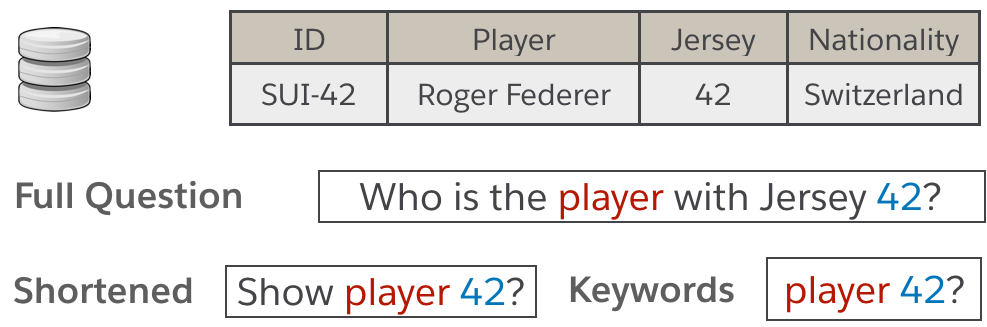}
    \caption{Examples of search-style user queries.}
    \label{fig:example}
\end{figure}

Prior research has primarily focused on translating grammatical, complete sentences to queries.
However, an internal user survey on the search service of a major customer relationship management (CRM) platform\footnote{\url{https://www.salesforce.com/}}
revealed that users have a tendency to communicate in a 
colloquial form which could vary from using only keywords (``player 42'') to very short phrases (``show player 42'') to complete sentences (``Who is the player who wears Jersey 42?''). Apart from variation in style, users dropping content words from their searches in the interest of brevity also has the potential consequence of making their questions ambiguous. This could render the task unsolvable even to models accustomed to the NLS style of text. For example, in Figure \ref{fig:example}, without the word ``Jersey'', it is impossible to identify which column's value (\texttt{Id} or \texttt{Jersey}) must equal 42.
 
In this work, we show that Text2SQL systems trained on only complete sentences struggle to adapt to the noisy keyword/short phrasal style of questions. To combat this, we introduce different data augmentation strategies inspired from our user search patterns and style.
\hide{
\todo{We have enough space, can you show 2-3 examples in this figure? No need to use a boxed layout, you can show a table. Also please use .pdf to save the figure, which has higher resolution.}
}
To tackle the induced ambiguity, a potential solution is to utilize the table content by allowing the model to scan the table for different terms present in the question and utilize that information to disambiguate (If the token ``42'' was only found in the \texttt{Jersey} column, then \texttt{Jersey} must be the column equal to 42).
Though effective, this approach could become prohibitively expensive (in terms of inference time or memory required) on large tables as the model would have to search over the entire of the table content for every question.

We hypothesize that in most cases, the model only needs samples from the table content and not the exact rows that match tokens in the NLS question to disambiguate columns. For example, if the \texttt{Id} column contained alpha-numeric IDs, \texttt{Player} and \texttt{Nationality} contained strings, and \texttt{Jersey} contained two digit numbers, then \texttt{Jersey} must be the column equal to 42.
Sampling alleviates the need of a full table scan for every question. The samples for each column could be generated offline and remain unchanged across questions or periodically refreshed (to reflect potential distribution shifts in the table or user queries), allowing for adaptation and personalization without retraining the model.

In summary, our contributions are as follows:
\begin{enumerate}
    \itemsep0em 
    \item We augment the well-formed WikiSQL dataset with synthetic search-style questions to adapt to short, colloquial input.
    \item We propose new models which incorporate table content in a BERT encoder via two sampling strategies to handle ambiguous questions. 
    \item We perform an in-depth qualitative and quantitative (accuracy, inference time, memory) analysis to show the efficacy of each content sampling strategy.
    \item We curate a dataset of 400 questions to benchmark performance of Text-to-SQL models in this setting.
\end{enumerate}

\hide{
\begin{enumerate}
    \item augmenting well-formed text with synthetic data to ensure robustness over NLS style questions 
    \item incorporating table content to a BERT model via two sampling strategies to handle ambiguous questions 
    \item an in-depth qualitative and quantitative (accuracy, inference time, memory) analysis showcasing the efficacy of both content sampling strategies 
    \item a curated dataset to benchmark performance on short, ambiguous questions.
\end{enumerate}
}

Apart from adapting to NLS style questions, \Ours also achieves state-of-the-art performance on the original WikiSQL~\cite{wikisql} dataset, 
outperforming all baselines that do not use execution guided decoding. We base our work off SQLova \cite{sqlova} but our methods are 
generalizable to other approaches\footnote{Our code and annotated data can be found at \\ \url{https://github.com/karthikradhakrishnan96/ColloQL}.}. 
\section{Related Work}
\paragraph{Text-to-SQL approaches for the WikiSQL benchmark} Text-to-SQL falls under a broader class of semantic parsing tasks and has been widely studied in the NLP and database communities. While early works have focused on pattern-matching and rule-based techniques~\cite{DBLP:journals/corr/cmp-lg-9503016,li2014nalir,setlur2016eviza
}, with the introduction of large scale datasets such as WikiSQL~\cite{wikisql}, recent works have focused on neural methods for generating SQL.  They can be broadly categorized into a few themes - sequence to sequence (Seq2Seq), sequence to tree (Seq2Tree), and SQL-Sketch (logical form) methods. 

Seq2Seq models frame the task as an encoder-decoder problem by trying to generate the SQL query token-by-token from the input question. However, as noted by~\citet{sqlnet} these models suffer from the ``order matters" issue where the model is forced to match the ordering of the where clauses. \citet{wikisql} employ reinforcement learning based method to overcome this issue but the gains from this has been limited as noted in \citet{sqlnet}. 
Seq2Tree models generate the SQL query as an abstract syntax tree (AST) instead of a token sequence~\cite{DBLP:conf/acl/GuoZGXLLZ19,DBLP:conf/acl/WangSLPR20}. These approaches define a generation grammar for SQL and learn to output the action sequence for constructing the AST~\cite{DBLP:conf/emnlp/YinN18}. Seq2Tree approaches are widely adopted for benchmarks that contain complex SQL queries~\cite{spider} as the syntactic constraints they adopt are effective at pruning the output search space and capturing structural dependencies. However, they do not show much advantage 
on the WikiSQL benchmark where the SQL ASTs are largely flat.

\begin{figure}[h!]
    \centering
  \includegraphics[scale=0.36]{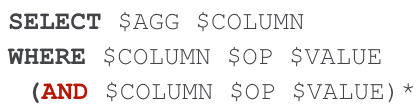}
  \caption{SQL-Sketch from \citet{sqlnet}.}
    \label{sketch}
\end{figure}

SQLNet \cite{sqlnet} introduces the concept of a SQL-Sketch, where it generates a sketch capturing the salient elements of the query as opposed to directly generating the query itself. SQLNet uses LSTMs to encode the question and headers and employs column attention to predict different components of the SQL-Sketch. As shown in Figure~\ref{sketch}, the query is decomposed into different components which are predicted individually. TypeSQL \cite{typesql} extends upon this approach by augmenting each token in the question with its type (whether it resembles the name of the column, FreeBase entity type, etc). SQLova \cite{sqlova} replaces the LSTMs encoder from SQLNet and uses BERT to encode the question and headers jointly. Unlike SQLNet, SQLova does not share any parameters in the decoders and identifies the where clause values using span detection instead of pointer generators. HydraNet~\cite{hydranet} breaks down the problem into column-wise ranking and decoding and assembles the outputs from each column to create the SQL query. 

\paragraph{Text-to-SQL with table content} Recent works like NL2SQL-RULE \cite{nl2sqlrule}, RAT-SQL~\cite{DBLP:conf/acl/WangSLPR20} and Photon \cite{photon} have looked into incorporating table content into the SQL generation. NL2SQL-RULE augments BERT representations 
with mark vectors for each question and table header token to indicate a match across the two parts. Photon only incorporates the content of a limited set of categorical fields when there is an exact match with a question token.
Unlike NL2SQL-RULE, ColloQL includes table content in the BERT encoder allowing it to form content-enhanced question and header representations and unlike Photon, ColloQL incorporates content for all columns and includes samples even when there is not an exact match to disambiguate columns effectively. TaBERT~\cite{DBLP:conf/acl/YinNYR20} lifted the idea further by pre-training joint representation of text and table taking into account row subsampled in a random or relevance-based manner. The pre-trained joint representation has been shown to outperform vanilla language models in several table QA and semantic parsing tasks.

\paragraph{Text-to-SQL with execution guided decoding} One common theme across all the high performing models on WikiSQL is that they all employ Execution Guided (EG) decoding. First introduced by~\citet{exec-guided},~EG is a technique where partial SQL queries are executed and their results are used to guide the decoding process. While EG has been shown to boost accuracy significantly, we do not apply execution guided decoding on our models for two reasons: 
Firstly, most EG methods modify the predicted query based on whether an empty set is returned. While this works well in the WikiSQL setting, having no results is often not due to an erroneous query. It is not uncommon for users to issue searches like ``my escalated support cases''(with the expectation of surfacing zero records) or ``John Doe leads''(to ensure that a record does not already exist before creating one) and we wanted to eliminate the reliance on database outputs to translate a query correctly. 
Secondly, database tables could have over 1M records and performing multiple database executions for every query could be expensive and is not always feasible whilst keeping up with the latency requirements of clients.

\paragraph{Text-to-SQL with noisy user input} While recent text-to-SQL research typically focus on benchmark datasets with complete and grammatical input, noisy user queries are commonly encountered in practical NLIDBs. Previous work have proposed several ways to address this issue.~\citet{zettlemoyer-collins-2007-online} introduced non-standard combinators to a combinatorial categorical grammar (GGG) based semantic parser to handle flexible word order and telegraphic language.~\citet{sajjad-etal-2012-underspecified} and~\citet{DBLP:conf/aaai/YaoLGSS19,DBLP:conf/emnlp/YaoSSY19} developed interactive semantic parsing models that generate clarification questions for user to complete their underspecified queries.~\citet{arthur-etal-2015-semantic} paraphrases an ambiguous input into a less ambiguous form.~\citet{10.1145/3301275.3302270} generates default logical forms for underspecified input.~\citet{photon} synthesized a new dataset and trained question filter to identify noisy user input and prompt user to rephrase.
Our work focus on handling short user utterances typically found in the search service of Salesforce CRM, where sampling-based content-aware models are effective at resolving most ambiguities. 
\section{Task and Datasets}
The Text2SQL task is to generate a SQL query from a natural language question and the database schema/content. In this work, we use the WikiSQL dataset \cite{wikisql} as it most closely matches the queries we expect to serve in a CRM.
Our users typically don't issue linguistically complex queries requiring joins or nesting but instead focus on filtering a single table based on certain clauses.

WikiSQL contains over 80K natural language questions distributed across 24K tables and their gold SQL queries. The performance is typically evaluated on two different types of accuracies - Logical Form (LF) and Execution (EX).
LF measures if the generated query exactly matches the gold query while EX executes the predicted and gold queries on the database and verifies if the answers returned by both are equal.
Note that LF is a stricter metric as many different SQL queries could produce the same output.

\begin{table*}[t]
\centering
\scalebox{0.9}{
\begin{tabular}{ll}
\toprule
\textbf{NL Search}    & \textbf{WikiSQL} \\ \midrule
acme opportunities    & which opportunities are for acme account  \\ 
John Doe accounts & where John Doe is owner what are accounts  \\ 
deals with revenue $>$10   & which deals have an expected revenue of over 10  \\ 
number of deals closed in 2019   & how many deals have closing year as 2019  \\ 
\bottomrule
\end{tabular}}
\caption{WikiSQL questions and their NLS-style counterparts.\label{tab:nlsvswikisql}}
\end{table*}

The WikiSQL dataset mostly comprises of verbose questions which differ in style as compared to the NLS questions issued by our users. Table \ref{tab:nlsvswikisql} shows NLS questions and their WikiSQL-style equivalents. To account for the differences in style, we augment the WikiSQL dataset with our synthetic data to simulate real-user NLS questions which is generated as follows.

\paragraph{Synthesizing user utterances from gold SQL labels}Since WikiSQL contains the gold labels for the SQL sketch, we can use this data to generate NLS-style questions. By analyzing our user search queries (which resemble those shown in Table \ref{tab:nlsvswikisql}) we built question templates which we fill based on the gold SQL-Sketch.
Some examples include shuffling the ordering of where conditions (users apply filters in different order), interchange ordering of column names and values (some users type ``US region cases'' while others type ``region US cases''), and insert the select column name in the beginning or the end of a question (``John Doe accounts'' vs ``accounts John Doe'').
The synthetic data is used in conjunction with clean well-formed queries from the original dataset, 
allowing the model to generalize to other queries not present in the templates. An example of synthetic utterances generated this way is shown below.

\textbf{Original Query} - Who is the player of Australian nationality that wears jersey number 42?

\textbf{Generated Queries} - \emph{player jersey 42 australian nationality; 42 jersey australian nationality player; australian nationality jersey 42 player; \ldots}

\paragraph{Supporting relational symbols in user utterance} We identify popular query ngrams when the conditional operator in the SQL-Sketch corresponds to either ``$>$'' or ``$<$'' and randomly replace these ngrams (``bigger than'', ``larger than'', etc) with the operator symbols, allowing our model to properly interpret them.

\paragraph{Controlled question simplification} Since WikiSQL contains no keyword-based questions and only a small portion of questions that are succinct enough to require reasoning over the table content, we employ a sentence simplification model followed by manual verification to create a test dataset to evaluate performance on NLS questions. A common user behavior is to drop unnecessary words from complete sentences to create shorter questions. We simulate this behavior by simplifying/compressing sentences to reduce verbosity. Note that keyword queries can be viewed as an extreme case of sentence simplification where only the required keywords are retained.

We make use of the controllable sentence simplifier by \citet{compression} to compress sentences to a desired length whilst retaining a specified set of keywords.
We specify the list of keywords to be the header name of the select column, the values in the where columns (we ignore the header names for the where columns as users tend to omit them from their queries).

In total, we create two datasets: \emph{short questions} with gold SQL labels and replacement of relation symbols, and \emph{simple questions} with controlled sentence simplification. 

\paragraph{Manually verified test set} We create a high-quality test set by manually verifying a subset of \emph{simple questions}\footnote{Sentence simplification creates a diverse set of examples which contains some of those generated by gold SQL label.}. 
A potential problem with sentence simplification models is ensuring that the shortened version still has enough information to execute the query correctly.
This could vary based on the table content and is difficult to identify if the query is impossible to be executed correctly.
We had a team of data scientists and engineers proficient in SQL to verify/correct outputs produced by the sentence simplification model and generated 400 queries 
for testing. We show examples in this dataset and report our manual quality evaluation in \S~\ref{sec:data-quality}.
\begin{figure*}[!h]
  \includegraphics[width=\textwidth]{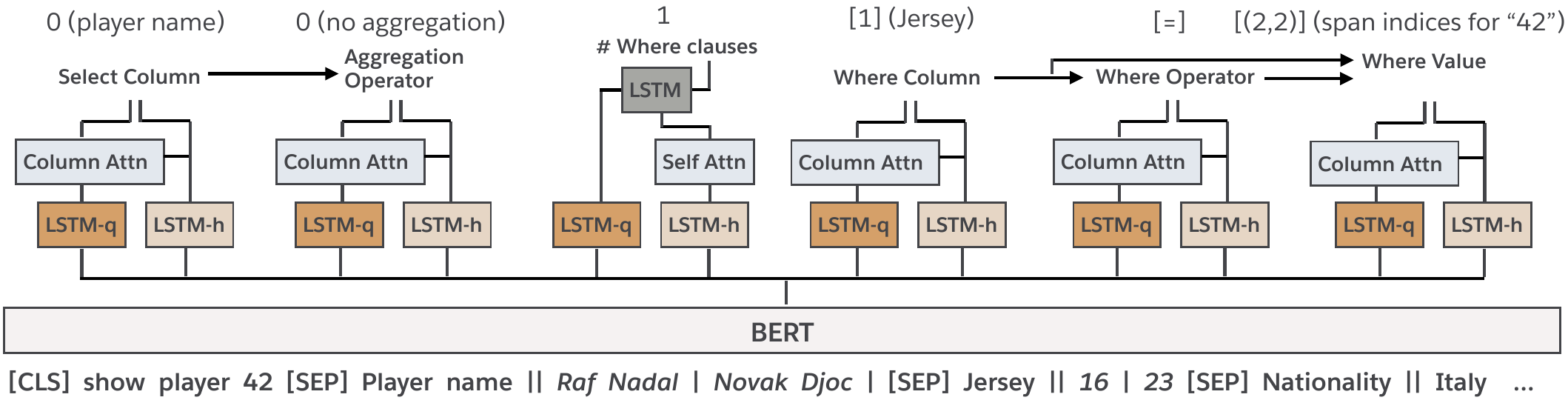}
  \caption{ColloQL uses the same NN architecture as SQLova where six decoding layers (one for each component of the SQL-Sketch) are used over BERT. The SQL query (\sqlquery{SELECT Player Name WHERE Jersey = 42}) is constructed from outputs of different components. Unlike SQLova, we also contextualize the question with the table samples (underlined in the figure) delimited by special tokens.}
  \label{arch}
\end{figure*}

\section{Proposed Approach}
Following \citet{sqlnet} and \citet{sqlova}, we decompose the SQL generation task into 6 different subtasks - one for each component of the SQL-Sketch. These subtasks all share a common encoder but use different decoder layers. The encoder is a BERT model \cite{devlin2018bert} which produces contextualized representations of the question, headers and the decoders largely use a task-specific LSTM with column-attention. Column-attention \cite{sqlnet} is a mechanism where each header attends over all query tokens to produce a single representation over which a dense layer is used to predict probabilities.

The select, aggregation, where-num, and where-operator branches use LSTMs + Column-attention followed by a softmax layer to output probabilities. The where-column branch is similar but uses a sigmoid instead as multiple columns could appear in the where clause and the where-value outputs start-end spans for the values from the question.

Figure \ref{arch} highlights the architecture of our model. We retain the same encoder-decoder architecture as SQLova as our main contribution lies in the data augmentation and content sampling techniques to handle NLS questions.

\subsection{Content Incorporation}
As highlighted previously, table content could be a useful feature in helping the model disambiguate between different columns. Consider a table of tennis players as shown below.
\begin{table}[h!]
\definecolor{header}{HTML}{DCD2CC}
\centering
\begin{tabular}{|l|l|l|}
\hline
\rowcolor{header}
\multicolumn{1}{c}{\textbf{Result}}    & \multicolumn{1}{c}{\textbf{Court}} & \multicolumn{1}{c}{\textbf{Player}}          \\ \hline

winner    & clay  & Rafael Nadal    \\ \hline
runner-up & grass & Novak Djokovic  \\ \hline
winner    & hard  & Jarkko Nieminen \\ \hline
\end{tabular}
\end{table}

Now, consider a question \textit{``courts with Rafael Nadal as winner''}. A model which isn't informed about the content of the table cannot easily understand that Rafael Nadal needs to be the where clause value for \texttt{Player} and winner for the \texttt{Result} column. Allowing the model to scan the table for entities like ``Rafael Nadal'' or ``winner'' could help the model incorporate table content effectively.

Consider another question \textit{``courts with Roger Federer as winner''}. It is intuitive that this query follows the same structure as the previous, except that the required value is now ``Roger Federer''. However, ``Roger Federer'' is not present in the table. We hypothesize that while table content is useful to the model, it does not need to be relevant to the query. The model, when given random samples of values for each column can infer the role of a particular column and generalize to unseen values which are similar to the column samples. In this work, we experiment with two sampling techniques - random and relevance sampling. 

\subsubsection{Random Sampling}
Random sampling uses a fixed set of question agnostic column values sampled randomly (without replacement) and does not require access to the table once the samples are created. Since the sampling process can be done entirely offline, it adds negligible memory and time to the query execution. Additionally, the model can now be used in privacy sensitive scenarios as it does not access the table content and the samples could be manually configured.  The model, now being content informed, performs better than its non-content counterparts whilst being more efficient than its full table content counterparts.

\subsubsection{Relevance Sampling}

Relevance sampling is used in cases where access to table is permitted and it includes a combination of samples relevant to question tokens and random samples. We index all cells of a table and perform a keyword search in the question to identify most relevant cells using FlashText~\cite{flashtext} and include them as samples. In situations where the number of keyword matches are fewer than intended for a column or there are no matches, we fallback on random sampling to select the remaining samples.

To illustrate the importance of including random samples in the relevance sampling strategy, consider the following example: 

\vspace{.1in}
\textbf{Question} - \emph{Which countries hosted the \textit{MHL} league?}

\texttt{League} \textbf{values} - NHL, MLB, NBA
\vspace{.1in}

Photon~\cite{photon}, a model which only includes up to a single matched value, interprets this query incorrectly (\sqlquery{Select country where league = MHL league}). Its value matching approach retrieves an empty set to augment the table.\footnote{We ran the evaluation on Photon's demo page.} 
Our model with relevance sampling tackles cases like this successfully (\sqlquery{Select country where league = MHL}) as NHL, MLB, and NBA were included as samples because of the fallback on random sampling. Including random samples improves the model's ability to interpret questions that have values not directly found in the table. 

The addition of random samples also allows the model to discriminate between columns effectively. Consider Question 4 from Table \ref{qual_table}, the question is ambiguous without table content because it is unclear if the column to be selected is \texttt{Place} or \texttt{Country}. The pattern ``where are\ldots from?'' indicates that the user's intent is to find a location and both column names seem like a reasonable choice (\texttt{Place} is a synonym for location and \texttt{Country} is a location). However, when augmented with random column samples, we see that the \texttt{Place} column only contains numeric values and is used as the synonym of ``rank'' in this table. 

Figure \ref{arch} shows our input representation to the BERT model. Our representation 
bears similarity to Photon where the content values are concatenated along with the headers and the question separated by special tokens. However, Photon only tackles columns with picklists (categorical columns storing small fixed set of values) while we support numeric and free-form text columns as well. Additionally, as mentioned above, since Photon only incorporates a single matched value, it doesn't gracefully interpret all questions.

We concatenate the column samples to the headers with special delimiters and experiment with 1,3,5 samples for each column. 
The number of samples is currently limited by the maximum sequence length supported by BERT models and in the future we hope to experiment with operating on each column individually~\cite{hydranet} and diversity based sampling to extract the most distinctive samples.
\section{Experiment Setup}
We use the base version of BERT in all our experiments and made necessary changes for sampling on the original SQLova codebase.
We use Adam~\cite{adam} optimizer with a learning rate of 1e-3 for the decoder layers and 1e-5 for the BERT model.

\section{Results}
\subsection{Qualitative Examples}

\begin{table*}[h!]

\begin{tabular}{p{1cm}p{14.1cm}}
\toprule 

\includegraphics[width=.14in]{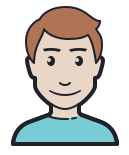} & grid of bmw rider with $>$ 200 laps                                                                                                                               \\ 
\includegraphics[width=.14in]{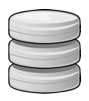}  & {Rider ${||}$ Nicolas Terol ${|}$ Mike Di Meglio ${|}$ Stevie Bonsey [SEP] Manufacturer ${||}$ Derbi ${|}$ Honda ${|}$ KTM [SEP] Laps ${||}$ 1 ${|}$ 24 ${|}$ 0 [SEP] Grid ${||}$ 20 ${|}$ 29 ${|}$ 25 \ldots} \\ 
                  SQL      & \sqlquery{SELECT (Grid) FROM 2-14125739-3 WHERE Manufacturer = bmw AND Laps $>$ 200}                                                                       \\
    \hline
\includegraphics[width=.14in]{src/fig/user.png} &  grid of maria herrera rider with $<$ 200 laps                                                                                                                               \\ 
\includegraphics[width=.14in]{src/fig/db.png}  & {Rider ${||}$ Nicolas Terol ${|}$ Mike Di Meglio ${|}$ Stevie Bonsey [SEP] Manufacturer ${||}$ Derbi ${|}$ Honda ${|}$ KTM [SEP] Laps ${||}$ 1 ${|}$ 24 ${|}$ 0 [SEP] Grid ${||}$ 20 ${|}$ 29 ${|}$ 25 \ldots} \\ 
                  SQL      & \sqlquery{SELECT(Grid) FROM 2-14125739-3 WHERE Rider = maria herrera AND Laps $<$ 200}                                                                       \\ 
    \hline
\includegraphics[width=.14in]{src/fig/user.png} & fox tv series female\\ 
\includegraphics[width=.14in]{src/fig/db.png}  & {Animal Name ${||}$ Jack ${|}$ The Big Owl ${|}$ The Wild Boar [SEP] Species ${||}$ \textbf{Fox} ${|}$ Badger ${|}$ Boar [SEP] Books ${||}$ No ${|}$ Yes [SEP] Gender ${||}$ male ${|}$ \textbf{female} \ldots} \\ 
                  SQL      & \sqlquery{SELECT(TV Series) FROM 2-11206371-5 WHERE Species = fox AND Gender = female}                                                                    \\ \hline
\includegraphics[width=.14in]{src/fig/user.png} &  Where are Charlie Freedman/Eddie Fletcher from?                                                                                                                               \\ 
\includegraphics[width=.14in]{src/fig/db.png}  & {Place ${||}$ 7 ${|}$ 9 ${|}$ 1 [SEP] Rider ${||}$ \textbf{Charlie Freedman/Eddie Fletcher} ${|}$ Mick Horsepole/E \ldots [SEP] Country ${||}$ West Germany ${|}$ Switzerland ${|}$ United Kingdom [SEP] \ldots} \\ 
                  SQL      & \sqlquery{SELECT(Country) FROM 2-10301911-6 WHERE Rider = charlie freedman/eddie fletcher}                                                                       \\ 
\bottomrule
\end{tabular}
    \caption{Some qualitative examples from our random (1,2) and relevance (3,4) sampling models. Bold values in headers indicate a match in the question.}
    \label{qual_table}
\end{table*}

Table~\ref{qual_table} shows some qualitative examples from our model when augmented with 3 values included for each column. The first two examples are based on random sampling and the latter two are based on relevance sampling.
Our model is able to correctly resolve phrases such as 
``Maria Herrera'' and ``BMW'' to the right columns when the corresponding values were 
not seen during training or inference. 
Consider the first two examples with different modifiers of ``rider'', leveraging the sampled values, our model correctly matches  ``BMW'' to \texttt{Manufacturer} (column storing brand name like values) and ``Maria Herrera'' to \texttt{Rider} (column storing human name like values). 

\subsection{Effect of Random Sampling}
We show performance of our model evaluated on the original WikiSQL dev dataset under different sampling settings. Owing to the 512 token limit, we only sample upto 5 values per column in Table \ref{tab:sampling}. Modifying the architecture to operate on one column at a time (HydraNet) would allow us to use more samples. Our model performs significantly better than our base SQLova model and performs competitively with other larger models.

\begin{table}[h!]
\centering
        \begin{tabular}{@{}l r r@{}}
            \toprule
            \textbf{Model}    & \textbf{LF (dev)}  & \textbf{EX (dev)}    \\ 
            \midrule
            SQLova\textsubscript{\textsc{base}}  & 79.5 & 85.3      \\
            SQLova\textsubscript{\textsc{large}} & 81.6 & 87.2 \\
            HydraNet\textsubscript{\textsc{large}} \textsuperscript{*} & \textbf{83.6} & 89.1 \\
            \Ours\textsubscript{rand:1}\textsuperscript{\dag}& 82.0 & 87.6 \\
            \Ours\textsubscript{rand:3}\textsuperscript{\dag}& 83.3 & 89.1 \\
            \Ours\textsubscript{rand:5}\textsuperscript{\dag}& 83.5 & \textbf{89.3} \\
            \bottomrule
        \end{tabular}
        \caption{Model performance with different sampling settings. Rand:[1,3,5] uses random sampling. $\dag$ indicates that data augmentation is added.
        }
        \label{tab:sampling}
\end{table}

\makeatletter
\def\blfootnote{\xdef\@thefnmark{}\@footnotetext}
\makeatother
\blfootnote{* \space Due to unavailability of code, HydraNet numbers are only reported on datasets used in their paper}

\subsection{Effect of Relevance Sampling}
In addition to random sampling, we also provide results on a model that finds the most relevant samples to the question. In Table \ref{tab:relsample}, we compare our results with NL2SQL-RULE~\cite{nl2sqlrule} (uses entire table content) and EM:1 (including a single exactly matched value), the content incorporation strategy adopted by Photon \cite{photon}. Since WikiSQL does not distinguish categorical columns, we applied the exact match to all columns. 
Our model achieves 85.2\% logical form and 90.65\% execution accuracy on the original WikiSQL dataset outperforming all models without EG.

\begin{table}[h!]
\centering
    \begin{tabular}{@{}lrr@{}}
        \toprule
        \textbf{Model}    & \textbf{LF (dev)}  & \textbf{EX (dev)}   \\ \midrule
        NL2SQL\textsubscript{\textsc{base}} & 84.3 & 90.3 \\
        \Ours\textsubscript{em:1}\textsuperscript{\dag\ddag} & 82.5 & 88.2 \\ 
        \Ours\textsubscript{rand:3}\textsuperscript{\dag} & 83.3 & 89.1 \\ 
        \Ours\textsubscript{rel:3}\textsuperscript{\dag} & \textbf{85.2} & \textbf{90.6} \\ 
        
        \bottomrule
    \end{tabular}
    \hide{\begin{tablenotes}
        \small
        \item \ddag Photon's equivalent
    \end{tablenotes}}
\caption{Efficacy of different content incorporation strategies. Relevance sampling (with 3 samples) gives the best performance. \ddag denotes our implementation of Photon.}
\label{tab:relsample}
\end{table}

We also studied the memory and time footprint for indexing cells with increasing table sizes by benchmarking the performance of random and relevance sampling on very large tables. To simulate real-world data, we used IMDB movie database - a large-scale database with tables spanning over 7M rows containing movie metadata.

The random sampling method is agnostic to table size as samples are generated just once while the relevance sampling method scans the table to pick the best samples for each query. The results are shown in Table \ref{tab:benchmark}.

\begin{table}[h!]
    \centering
        \begin{tabular}{@{}lrrrr@{}}
            \toprule
            \textbf{Model}    & \textbf{Rows}  & \textbf{Exec} & \textbf{RAM} & \textbf{Setup}   \\ \midrule
            \Ours\textsubscript{rand:3} & 1M  & \textbf{0.2s} & - & -   \\ 
            \Ours\textsubscript{rel:3} & 1M  & 1.5s & 4G & 20s   \\ 
            NL2SQL\textsubscript{\textsc{base}} & 1M  & 200s & 1.4G & -   \\ \hdashline
            \Ours\textsubscript{rand:3} & 7M  & \textbf{0.2s} & - & -   \\ 
            \Ours\textsubscript{rel:3} & 7M  & 15s & 18G & 60s   \\ 
            NL2SQL\textsubscript{\textsc{base}} & 7M  & x & 8G & -   \\ \bottomrule
        \end{tabular}
        \begin{tablenotes}
            \small
            \item `-' $\rightarrow$ negligible; `x' $\rightarrow$ practically intractable
        \end{tablenotes}
        \caption{Benchmarking different content incorporation strategies with respect to execution time (CPU), memory footprint and setup time (for indexing).}
        \label{tab:benchmark}
\end{table}

\subsection{Performance on Simple Questions}

To measure the efficacy of content augmentation, we compared \Ours with other works on our dataset of 400 simplified queries which was generated by the sentence simplification model and verified/corrected by a team of data scientists and engineers. This dataset largely contains queries in which the where columns are not explicitly mentioned in the query and requires the model to infer them. We can see from Table \ref{tab:simplified} that a model un-informed of the content drops in accuracy (especially in the where column prediction) while \Ours retains its performance.

\begin{table}[h!]
\centering
\begin{tabular}{@{}lrr@{}}
\toprule
\textbf{Model}    & \textbf{LF}  & \textbf{Where-col acc}    \\ \midrule
SQLova\textsubscript{\textsc{base}} & 68.7 & 78.2 \\ 
NL2SQL\textsubscript{\textsc{base}} & 80.8 & 94.3 \\ 
\Ours\textsubscript{rand:5}\textsuperscript{\dag} & 83.2 & 92.2 \\ 
\Ours\textsubscript{rel:3}\textsuperscript{\dag} & \textbf{87.0} & \textbf{97.2} \\ \bottomrule
\end{tabular}
\caption{Performance on the curated test set i.e. 400 simplified queries.}
\label{tab:simplified}
\end{table}

\subsection{Effect of Augmentation}

Since SQLova was originally trained with complete sentences, it does not adapt well to short questions. Retraining the same model with augmented data from our templates recovers the performance (tested using short questions). Additionally, the augmentation also results in improved generalization resulting in a minor LF accuracy improvement on the original dev data as shown in Table \ref{tab:augmentation}.

\begin{table}[h!]
\centering
\begin{tabular}{@{}l r r@{}}
\toprule
\textbf{Model}    & \textbf{LF(short)} & \textbf{LF(dev)}     \\
\midrule
SQLova\textsubscript{\textsc{base}}  & 54.0 & 79.5      \\
SQLova\textsubscript{\textsc{base}} & 86.2  & 80.6\\ 
\bottomrule
\end{tabular}
\caption{Comparing logical form accuracy of SQLova with augmentation. LF(short) is the dev accuracy on the short questions. LF(dev) is the accuracy on the WikiSQL dev split.}
\label{tab:augmentation}
\end{table}


\subsection{Performance on WikiSQL test set}
Finally, we also show the performance of our model on the WikiSQL test dataset comparing them to the top approaches on the WikiSQL leaderboard\footnote{\url{https://github.com/salesforce/WikiSQL}}. As we can see in Table \ref{tab:testset}, \Ours achieves the highest accuracy without execution guided decoding on the WikiSQL test set.

\begin{table}[h!]
\centering
\begin{tabular}{@{}lrr@{}}
\toprule
\textbf{Model}    & \textbf{LF(test)}  & \textbf{EX(test)}    \\ \midrule
HydraNet\textsubscript{\textsc{large}} & 83.8 & 89.2 \\ 
NL2SQL\textsubscript{\textsc{base}} & 83.7 & 89.2 \\ 
\Ours\textsubscript{rel:3}\textsuperscript{\dag} & \textbf{84.9} & \textbf{90.7} \\ \bottomrule
\end{tabular}
\caption{Performance on the WikiSQL test set.}
\label{tab:testset}
\end{table}
\section{Error Analysis}

We classified the errors made by our model on the ColloQL curated dataset into two major categories:

\textbf{Aggregation} - Given that WikiSQL contains noisy labels for aggregation component \cite{sqlova} and the model was optimized for accuracy on WikiSQL, there are some errors in predicting this slot.

\textbf{Select Columns} - The simplified questions 
are often more ambiguous when predicting whether a column is a target to be selected or is used in a filtering condition (e.g. for the question ``smallest tiesplayed 6 years'', the model interprets it as \sqlquery{SELECT MIN(years) WHERE tiesplayed = 6} while the correct query is \sqlquery{SELECT MIN(tiesplayed) WHERE years = 6}). Additionally, we noticed that our annotators simplified column headers like ``shortstop'' and ``rightfielder'' to ``SS'' and ``RF'', making the question very difficult to solve. 
\section{Conclusion and Future Work}

In this work we tackled the task of converting noisy (short, potentially ambiguous) search-like (NLS) questions to SQL queries. We introduced data augmentation strategies to adapt to the NLS style of text and a novel content enhancement to BERT via two sampling strategies - random and relevance sampling. Random sampling overcomes some of the performance / privacy challenges of incorporating table content and relevance sampling achieves state-of-the-art performance when access to table content is permitted. Finally, we also curated a new held-out dataset to evaluate performance against NLS questions. 

In the future, we hope to explore different sampling techniques (based on user history, sampling to maximize discernment between columns) to enhance performance. Besides, our approach and dataset mainly target telegraphic queries that can be effectively disambiguated with table contents, which frequency occur in our search service. We plan to extend our work to handle other types of input ambiguities and other application domains. 

\section*{Acknowledgments}
We would like to thank Christian Posse and Mario Rodriguez for their support, help and invaluable feedback throughout the development of this work. We also would like to thank our team of expert annotators for their contribution.

\bibliography{main}

\begin{thebibliography}{27}
\expandafter\ifx\csname natexlab\endcsname\relax\def\natexlab#1{#1}\fi

\bibitem[{Androutsopoulos et~al.(1995)Androutsopoulos, Ritchie, and
  Thanisch}]{DBLP:journals/corr/cmp-lg-9503016}
Ion Androutsopoulos, Graeme~D. Ritchie, and Peter Thanisch. 1995.
\newblock \href {http://arxiv.org/abs/cmp-lg/9503016} {Natural language
  interfaces to databases - an introduction}.
\newblock \emph{CoRR}, cmp-lg/9503016.

\bibitem[{Arthur et~al.(2015)Arthur, Neubig, Sakti, Toda, and
  Nakamura}]{arthur-etal-2015-semantic}
Philip Arthur, Graham Neubig, Sakriani Sakti, Tomoki Toda, and Satoshi
  Nakamura. 2015.
\newblock \href {https://doi.org/10.1162/tacl_a_00159} {Semantic parsing of
  ambiguous input through paraphrasing and verification}.
\newblock \emph{Transactions of the Association for Computational Linguistics},
  3:571--584.

\bibitem[{Devlin et~al.(2018)Devlin, Chang, Lee, and
  Toutanova}]{devlin2018bert}
Jacob Devlin, Ming-Wei Chang, Kenton Lee, and Kristina Toutanova. 2018.
\newblock Bert: Pre-training of deep bidirectional transformers for language
  understanding.
\newblock \emph{arXiv preprint arXiv:1810.04805}.

\bibitem[{Finegan{-}Dollak et~al.(2018)Finegan{-}Dollak, Kummerfeld, Zhang,
  Ramanathan, Sadasivam, Zhang, and Radev}]{DBLP:conf/acl/RadevKZZFRS18}
Catherine Finegan{-}Dollak, Jonathan~K. Kummerfeld, Li~Zhang, Karthik
  Ramanathan, Sesh Sadasivam, Rui Zhang, and Dragomir~R. Radev. 2018.
\newblock \href {https://doi.org/10.18653/v1/P18-1033} {Improving text-to-sql
  evaluation methodology}.
\newblock In \emph{Proceedings of the 56th Annual Meeting of the Association
  for Computational Linguistics, {ACL} 2018, Melbourne, Australia, July 15-20,
  2018, Volume 1: Long Papers}, pages 351--360. Association for Computational
  Linguistics.

\bibitem[{Guo et~al.(2019)Guo, Zhan, Gao, Xiao, Lou, Liu, and
  Zhang}]{DBLP:conf/acl/GuoZGXLLZ19}
Jiaqi Guo, Zecheng Zhan, Yan Gao, Yan Xiao, Jian{-}Guang Lou, Ting Liu, and
  Dongmei Zhang. 2019.
\newblock \href {https://doi.org/10.18653/v1/p19-1444} {Towards complex
  text-to-sql in cross-domain database with intermediate representation}.
\newblock In \emph{Proceedings of the 57th Conference of the Association for
  Computational Linguistics, {ACL} 2019, Florence, Italy, July 28- August 2,
  2019, Volume 1: Long Papers}, pages 4524--4535. Association for Computational
  Linguistics.

\bibitem[{Guo and Gao(2019)}]{nl2sqlrule}
Tong Guo and Huilin Gao. 2019.
\newblock Content enhanced bert-based text-to-sql generation.
\newblock \emph{arXiv preprint arXiv:1910.07179}.

\bibitem[{Handler and O{'}Connor(2019)}]{compression}
Abram Handler and Brendan O{'}Connor. 2019.
\newblock \href {https://doi.org/10.18653/v1/D19-1612} {Query-focused sentence
  compression in linear time}.
\newblock In \emph{Proceedings of the 2019 Conference on Empirical Methods in
  Natural Language Processing and the 9th International Joint Conference on
  Natural Language Processing (EMNLP-IJCNLP)}, pages 5969--5975, Hong Kong,
  China. Association for Computational Linguistics.

\bibitem[{Hwang et~al.(2019)Hwang, Yim, Park, and Seo}]{sqlova}
Wonseok Hwang, Jinyeung Yim, Seunghyun Park, and Minjoon Seo. 2019.
\newblock A comprehensive exploration on wikisql with table-aware word
  contextualization.
\newblock \emph{ArXiv}, abs/1902.01069.

\bibitem[{Kingma and Ba(2019)}]{adam}
Diederik~P Kingma and J~Adam Ba. 2019.
\newblock A method for stochastic optimization. arxiv 2014.
\newblock \emph{arXiv preprint arXiv:1412.6980}, 434.

\bibitem[{Li and Jagadish(2014)}]{li2014nalir}
Fei Li and Hosagrahar~V Jagadish. 2014.
\newblock Nalir: an interactive natural language interface for querying
  relational databases.
\newblock In \emph{Proceedings of the 2014 ACM SIGMOD international conference
  on Management of data}, pages 709--712.

\bibitem[{Lyu et~al.(2020)Lyu, Chakrabarti, Hathi, Kundu, Zhang, and
  Chen}]{hydranet}
Qin Lyu, Kaushik Chakrabarti, Shobhit Hathi, Souvik Kundu, Jianwen Zhang, and
  Zheng Chen. 2020.
\newblock \href
  {https://www.microsoft.com/en-us/research/publication/hybrid-ranking-network-for-text-to-sql/}
  {Hybrid ranking network for text-to-sql}.
\newblock Technical Report MSR-TR-2020-7, Microsoft Dynamics 365 AI.

\bibitem[{Sajjad et~al.(2012)Sajjad, Pantel, and
  Gamon}]{sajjad-etal-2012-underspecified}
Hassan Sajjad, Patrick Pantel, and Michael Gamon. 2012.
\newblock \href {https://www.aclweb.org/anthology/C12-1143} {Underspecified
  query refinement via natural language question generation}.
\newblock In \emph{Proceedings of {COLING} 2012}, pages 2341--2356, Mumbai,
  India. The COLING 2012 Organizing Committee.

\bibitem[{Setlur et~al.(2016)Setlur, Battersby, Tory, Gossweiler, and
  Chang}]{setlur2016eviza}
Vidya Setlur, Sarah~E Battersby, Melanie Tory, Rich Gossweiler, and Angel~X
  Chang. 2016.
\newblock Eviza: A natural language interface for visual analysis.
\newblock In \emph{Proceedings of the 29th Annual Symposium on User Interface
  Software and Technology}, pages 365--377.

\bibitem[{Setlur et~al.(2019)Setlur, Tory, and
  Djalali}]{10.1145/3301275.3302270}
Vidya Setlur, Melanie Tory, and Alex Djalali. 2019.
\newblock \href {https://doi.org/10.1145/3301275.3302270} {Inferencing
  underspecified natural language utterances in visual analysis}.
\newblock In \emph{Proceedings of the 24th International Conference on
  Intelligent User Interfaces}, IUI '19, page 40–51, New York, NY, USA.
  Association for Computing Machinery.

\bibitem[{{Singh}(2017)}]{flashtext}
V.~{Singh}. 2017.
\newblock \href {http://arxiv.org/abs/1711.00046} {{Replace or Retrieve
  Keywords In Documents at Scale}}.
\newblock \emph{ArXiv e-prints}.

\bibitem[{Wang et~al.(2020)Wang, Shin, Liu, Polozov, and
  Richardson}]{DBLP:conf/acl/WangSLPR20}
Bailin Wang, Richard Shin, Xiaodong Liu, Oleksandr Polozov, and Matthew
  Richardson. 2020.
\newblock \href {https://www.aclweb.org/anthology/2020.acl-main.677/}
  {{RAT-SQL:} relation-aware schema encoding and linking for text-to-sql
  parsers}.
\newblock In \emph{Proceedings of the 58th Annual Meeting of the Association
  for Computational Linguistics, {ACL} 2020, Online, July 5-10, 2020}, pages
  7567--7578. Association for Computational Linguistics.

\bibitem[{Wang et~al.(2018)Wang, Tatwawadi, Brockschmidt, Huang, Mao, Polozov,
  and Singh}]{exec-guided}
Chenglong Wang, Kedar Tatwawadi, Marc Brockschmidt, Po-Sen Huang, Yi~Mao,
  Oleksandr Polozov, and Rishabh Singh. 2018.
\newblock Robust text-to-sql generation with execution-guided decoding.
\newblock \emph{arXiv preprint arXiv:1807.03100}.

\bibitem[{Xu et~al.(2018)Xu, Liu, and Song}]{sqlnet}
Xiaojun Xu, Chang Liu, and Dawn Song. 2018.
\newblock Sqlnet: Generating structured queries from natural language without
  reinforcement learning.

\bibitem[{Yao et~al.(2019{\natexlab{a}})Yao, Li, Gao, Sadler, and
  Sun}]{DBLP:conf/aaai/YaoLGSS19}
Ziyu Yao, Xiujun Li, Jianfeng Gao, Brian~M. Sadler, and Huan Sun.
  2019{\natexlab{a}}.
\newblock \href {https://doi.org/10.1609/aaai.v33i01.33012547} {Interactive
  semantic parsing for if-then recipes via hierarchical reinforcement
  learning}.
\newblock In \emph{The Thirty-Third {AAAI} Conference on Artificial
  Intelligence, {AAAI} 2019, The Thirty-First Innovative Applications of
  Artificial Intelligence Conference, {IAAI} 2019, The Ninth {AAAI} Symposium
  on Educational Advances in Artificial Intelligence, {EAAI} 2019, Honolulu,
  Hawaii, USA, January 27 - February 1, 2019}, pages 2547--2554. {AAAI} Press.

\bibitem[{Yao et~al.(2019{\natexlab{b}})Yao, Su, Sun, and
  Yih}]{DBLP:conf/emnlp/YaoSSY19}
Ziyu Yao, Yu~Su, Huan Sun, and Wen{-}tau Yih. 2019{\natexlab{b}}.
\newblock \href {https://doi.org/10.18653/v1/D19-1547} {Model-based interactive
  semantic parsing: {A} unified framework and {A} text-to-sql case study}.
\newblock In \emph{Proceedings of the 2019 Conference on Empirical Methods in
  Natural Language Processing and the 9th International Joint Conference on
  Natural Language Processing, {EMNLP-IJCNLP} 2019, Hong Kong, China, November
  3-7, 2019}, pages 5446--5457. Association for Computational Linguistics.

\bibitem[{Yin and Neubig(2018)}]{DBLP:conf/emnlp/YinN18}
Pengcheng Yin and Graham Neubig. 2018.
\newblock \href {https://doi.org/10.18653/v1/d18-2002} {{TRANX:} {A}
  transition-based neural abstract syntax parser for semantic parsing and code
  generation}.
\newblock In \emph{Proceedings of the 2018 Conference on Empirical Methods in
  Natural Language Processing, {EMNLP} 2018: System Demonstrations, Brussels,
  Belgium, October 31 - November 4, 2018}, pages 7--12. Association for
  Computational Linguistics.

\bibitem[{Yin et~al.(2020)Yin, Neubig, Yih, and
  Riedel}]{DBLP:conf/acl/YinNYR20}
Pengcheng Yin, Graham Neubig, Wen{-}tau Yih, and Sebastian Riedel. 2020.
\newblock \href {https://www.aclweb.org/anthology/2020.acl-main.745/} {Tabert:
  Pretraining for joint understanding of textual and tabular data}.
\newblock In \emph{Proceedings of the 58th Annual Meeting of the Association
  for Computational Linguistics, {ACL} 2020, Online, July 5-10, 2020}, pages
  8413--8426. Association for Computational Linguistics.

\bibitem[{Yu et~al.(2018{\natexlab{a}})Yu, Li, Zhang, Zhang, and
  Radev}]{typesql}
Tao Yu, Zifan Li, Zilin Zhang, Rui Zhang, and Dragomir Radev.
  2018{\natexlab{a}}.
\newblock Typesql: Knowledge-based type-aware neural text-to-sql generation.
\newblock \emph{arXiv preprint arXiv:1804.09769}.

\bibitem[{Yu et~al.(2018{\natexlab{b}})Yu, Zhang, Yang, Yasunaga, Wang, Li, Ma,
  Li, Yao, Roman, Zhang, and Radev}]{spider}
Tao Yu, Rui Zhang, Kai Yang, Michihiro Yasunaga, Dongxu Wang, Zifan Li, James
  Ma, Irene Li, Qingning Yao, Shanelle Roman, Zilin Zhang, and Dragomir Radev.
  2018{\natexlab{b}}.
\newblock Spider: A large-scale human-labeled dataset for complex and
  cross-domain semantic parsing and text-to-sql task.
\newblock In \emph{Proceedings of the 2018 Conference on Empirical Methods in
  Natural Language Processing}, Brussels, Belgium. Association for
  Computational Linguistics.

\bibitem[{Zeng et~al.(2020)Zeng, Lin, Hoi, Socher, Xiong, Lyu, and
  King}]{photon}
Jichuan Zeng, Xi~Victoria Lin, Steven~C.H. Hoi, Richard Socher, Caiming Xiong,
  Michael Lyu, and Irwin King. 2020.
\newblock \href {https://doi.org/10.18653/v1/2020.acl-demos.24} {{P}hoton: A
  robust cross-domain text-to-{SQL} system}.
\newblock In \emph{Proceedings of the 58th Annual Meeting of the Association
  for Computational Linguistics: System Demonstrations}, pages 204--214,
  Online. Association for Computational Linguistics.

\bibitem[{Zettlemoyer and Collins(2007)}]{zettlemoyer-collins-2007-online}
Luke Zettlemoyer and Michael Collins. 2007.
\newblock \href {https://www.aclweb.org/anthology/D07-1071} {Online learning of
  relaxed {CCG} grammars for parsing to logical form}.
\newblock In \emph{Proceedings of the 2007 Joint Conference on Empirical
  Methods in Natural Language Processing and Computational Natural Language
  Learning ({EMNLP}-{C}o{NLL})}, pages 678--687, Prague, Czech Republic.
  Association for Computational Linguistics.

\bibitem[{Zhong et~al.(2017)Zhong, Xiong, and Socher}]{wikisql}
Victor Zhong, Caiming Xiong, and Richard Socher. 2017.
\newblock Seq2sql: Generating structured queries from natural language using
  reinforcement learning.
\newblock \emph{CoRR}, abs/1709.00103.

\end{thebibliography}
\bibliographystyle{acl_natbib}

\newpage
\appendix

\section{Appendix}

\subsection{Test Set Quality}
\label{sec:data-quality}

One of the authors who did not participate in the dataset annotation randomly sampled 16/400 examples and manually checked the quality. 4/16 annotations were found to have issues in the natural language annotation. 

Table~\ref{tab:dataset_sample} shows examples from the \emph{simple question} dataset. The first 4 examples are correct, high-quality annotations while the bottom 4 are those with issues found during manual check. The high-quality simple question annotations are readable and on average have a smaller compression ratio compared to the noisy annotations. 

We noticed that some errors in the WikiSQL annotation~\cite{sqlova} were corrected when the simplified questions were produced, but some perpetuated through. In the second example, the annotator corrected spelling errors in the original WikiSQL annotation. 
However, in the 7th example, the original question misinterpreted \texttt{Year acquired} as a quantity and our simplified question inherited that error. Similarly, in the 8th example, the original question misinterpreted the field \texttt{Finalists} as ``score'' (it should represent ``number of finalists'') and our simplified question inherited it.

The 5th and 6th examples have unreadable questions as a result of sentence simplification (but our annotators still labeled them as correct). This is an artifact of the dataset as such unreadable, keyword-style queries may favor models that leverage table content to identify the columns. On the other hand, such queries could be useful as being able to interpret them may give users more flexibility when searching the content of a database.

\begin{table*}[h!]
\centering
\scalebox{.95}{
\begin{tabular}{p{1.7cm}p{13.6cm}}
\toprule 
Original& What is the amount of trees, that require replacement when the district is motovilikhinsky? \\ 
Simple & the amount of trees, that require replacement district motovilikhinsky? \\ 
\includegraphics[width=.14in]{src/fig/user.png}  & {District ${||}$ Total amount of trees ${||}$  Prevailing types, \% ${||}$ Amount of old trees ${||}$ Amount of trees, that require replacement} ${||}$ ...\\ 
                  SQL      & \sqlquery{SELECT (Amount of trees, that require replacement) from \_ WHERE District=Leninsky}                                                                    \\ \hline
Original &  How many winning drivers were the for the rnd equalling 5?                                                                                                                               \\ 
Simple &  how many winning drivers for 5?                                                                                                                               \\ 
\includegraphics[width=.14in]{src/fig/user.png}  & {Rnd ${||}$ Race Name ${||}$ Circuit ${||}$ City/Location ${||}$ Date ${||}$ Pole position ${||}$ Winning driver ${||}$ ...} \\ 
                  SQL      & \sqlquery{SELECT COUNT(Winning driver) from \_ WHERE Rnd=5}                                                                       \\ 
\hline
Original& For the episode(s) aired in the U.S. on 4 april 2008, what were the names?\\ 
Simple & for the episode(s) aired in U.S. 4 april 2008, names? \\ 
\includegraphics[width=.14in]{src/fig/user.png}  & {No. in season ${||}$ No. in series ${||}$ Title ${||}$ Canadian airdate ${||}$ US airdate ${||}$ Production code \ldots} \\ 
                  SQL      & \sqlquery{SELECT (Title) from \_ WHERE US airdate=4 April 2008}                                                                    \\ \hline
Original &  List the scores of all games when Miami were listed as the first Semi finalist?                                                                                                                               \\ 
Simple &  scores with miami listed as first semi finalist?                                                                                                                               \\ 
\includegraphics[width=.14in]{src/fig/user.png}  & {Year ${||}$ Champion ${||}$ Score ${||}$ Runner-Up ${||}$ Location ${||}$ Semi-Finalist \#1 ${||}$ Semi-Finalist \#2 \ldots} \\ 
                  SQL      & \sqlquery{SELECT (Score) from \_ WHERE Semi-Finalist \#1=Miami}                                                                       \\ 
    \hline\hline
Original& What school did the forward whose number is 10 belong to?\\ 
Simple & what school did forward 10 \\ 
\includegraphics[width=.14in]{src/fig/user.png}  & {Player ${||}$ No.(s) ${||}$ Height in Ft. ${||}$ Position ${||}$ Years for Rockets ${||}$ School/Club Team/Country \ldots} \\ 
                  SQL      & \sqlquery{SELECT (School/Club Team/Country) from \_ WHERE No.(s)=10 AND Position=Forward}                                                                    \\ \hline
Original &  Which visitors have a leading scorer of roy : 25?                                                                                                                               \\ 
Simple &  visitor roy : 25                                                                                                                            \\ 
\includegraphics[width=.14in]{src/fig/user.png}  & {\# ${||}$ Date ${||}$ Visitor ${||}$ Score ${||}$ Home ${||}$ Leading scorer ${||}$ Attendance ${||}$ Record ${||}$ Streak \ldots} \\ 
                  SQL      & \sqlquery{SELECT (Visitor) from \_ WHERE Leading scorer=Roy : 25}                                                                       \\ 
\hline
Original& how any were gained as the chan\\ 
Simple & how many gained chan\\ 
\includegraphics[width=.14in]{src/fig/user.png}  & { City ${||}$ Station ${||}$ Year acquired ${||}$ Primary programming source ${||}$ Other programming sources \ldots} \\ 
                  SQL      & \sqlquery{SELECT COUNT(Year acquired) from \_ WHERE Station=CHAN}                                                                    \\ \hline
Original &  What are the names that had a finalist score of 2??                                                                                                                               \\ 
Simple &  names that had finalist score 2?                                                                                                                               \\ 
\includegraphics[width=.14in]{src/fig/user.png}  & {School ${||}$ Winners ${||}$ Finalists ${||}$ Total Finals ${||}$ Year of last win} \\ 
                  SQL      & \sqlquery{SELECT (School) from \_ WHERE Finalists=2}                                                                       \\ 
\bottomrule
\end{tabular}
}
    \caption{Examples in \emph{simple questions} dev set. We use ``\_'' as placeholder for table in the SQL queries. Only table headers were shown. The top 4 examples are correct while the bottom 4 have issue in the natural language annotation.}
    \label{tab:dataset_sample}
\end{table*}

\end{document}